\documentclass[11pt]{article}

\usepackage[preprint]{acl}

\usepackage[ruled,vlined,linesnumbered]{algorithm2e}

\newcommand{\algstage}[1]{%
\vspace{2mm}
\hrule
\vspace{1mm}
\centerline{\large\bfseries #1}
\vspace{1mm}
\hrule
\vspace{2mm}
}

\usepackage{times}
\usepackage{latexsym}
\usepackage{booktabs}

\usepackage[T1]{fontenc}

\usepackage{amsmath}

\usepackage[utf8]{inputenc}
\usepackage{array}

\usepackage{microtype}
\usepackage[dvipsnames]{xcolor}
\usepackage{caption}

\usepackage{inconsolata}

\usepackage{graphicx}

\usepackage{tikz}
\usepackage[most]{tcolorbox}
\usepackage{enumitem}
\usetikzlibrary{fit, backgrounds, shapes.geometric}

\tcbset{
  narrativebox/.style={
    enhanced,
    colback=white,
    colframe=black!25,
    coltitle=black,
    colbacktitle=black!10,
    fonttitle=\bfseries\footnotesize,
    boxrule=0.35pt,
    arc=1.2mm,
    left=1.5mm,
    right=1.5mm,
    top=1.1mm,
    bottom=1.1mm,
    title filled,
    height=3.7cm,
    valign=center,
    before skip=4pt,
    after skip=4pt
  }
}

\title{\textsc{CART}: Clustering-based Analysis of Recurring narratives

for Taxonomy construction}

\author{
  \textbf{Aggelos Semoglou\textsuperscript{1,2}}
  \quad
  \textbf{Thomas Tselentis\textsuperscript{1}}
  \quad
  \textbf{John Pavlopoulos\textsuperscript{1,2}}
\\
\\
  \textsuperscript{1}Athens University of Economics and Business, Greece \\
  \textsuperscript{2}Archimedes Research Unit, Athena Research Center, Greece
\\
  \small{
    \textbf{Correspondence:}
    \href{mailto:a.semoglou@athenarc.gr}{a.semoglou@athenarc.gr}
  }
}

\begin{document}
\maketitle
\begin{abstract}
We present \textsc{CART}, a computational pipeline for extracting political narratives from parliamentary discourse and organizing them into a data-driven taxonomy without relying on predefined narrative categories. It encodes parliamentary discourse using semantic representations, filters relevant and reliable content through unsupervised confidence estimation, extracts narrative statements using LLMs, and uses narrative clustering and cluster analysis to construct a narrative taxonomy. Applied to Greek Parliament proceedings surrounding the 2023 Tempi train crash, it identifies ten recurring narrative families concerning responsibility attribution, institutional failure, railway safety, accountability, privatization, and the political interpretation of the tragedy. Our results demonstrate how combining unsupervised learning with LLM narrative extraction enables large parliamentary corpora to be transformed into interpretable, data-driven taxonomies of political narratives.
\end{abstract}

\section{Introduction}\label{sec:intro}
On February 28, 2023, a passenger train collided head-on with a freight train near Tempi, Greece, causing 57 deaths. The disaster prompted sustained public and parliamentary debate over its causes, institutional responsibility, and the subsequent investigation.\footnote{\href{https://www.hellenicparliament.gr/Koinovouleftikes-Epitropes/ektheseis?search=on&commissionType=59b8c6aa-ca8a-4da0-8b39-a190bcd7533d}{Preliminary Committees}, \href{https://www.hellenicparliament.gr/Koinovouleftikes-Epitropes/ektheseis?search=on&commissionType=caa37b07-7dc4-4b59-b17f-64c13915b29f}{Inquiry Committees}.}
These developments generated a large body of parliamentary discourse in which political actors framed the causes of the crash, assigned responsibility, and interpreted the investigation.

From a computational perspective, this setting poses a challenging narrative analysis problem. Parliamentary discourse is large, heterogeneous, and only partly relevant to the event, while no annotated relevance labels or predefined narrative categories are available. Political narratives must therefore be inferred from recurring interpretive claims. Manual coding is difficult at scale, making it necessary to identify relevant discourse, extract narratives, and organize them into interpretable families.

Political narratives shape policy~\cite{hajer1995politics}, provide causal interpretations and attribute responsibility~\cite{stone1989causal}, function as persuasive mechanisms~\cite{jones2014IntrNPF}, and enable emotional engagement~\cite{mclaughlin2019political}. Drawing on~\citet{nikolaidis2025polynarrative} and~\citet{patterson1998narrative}, we define political narratives as recurring, overt or implicit claims that promote a particular interpretation of an ongoing political issue. In this study, a \emph{narrative statement} is a concise claim about the Tempi crash, its causes, responsibility, the investigation, or its broader political interpretation, while a \emph{narrative family} is a group of semantically related statements expressing a recurring interpretation.

Advances in Natural Language Processing (NLP), Machine Learning (ML), and Large Language Models (LLMs) have supported growing interest in the analysis of narratives and political discourse~\cite{anning2023connecting, berta2024decoding, kotseva2023trend}. However, existing approaches typically address thematic analysis, narrative extraction, or taxonomy construction separately, providing limited support for settings in which relevant discourse must first be isolated from a much larger corpus and narrative categories are not known in advance. This motivates the need for a workflow that combines thematic filtering, narrative extraction, and taxonomy construction.

\begin{figure*}[t]
    \centering
    \includegraphics[width=\textwidth]{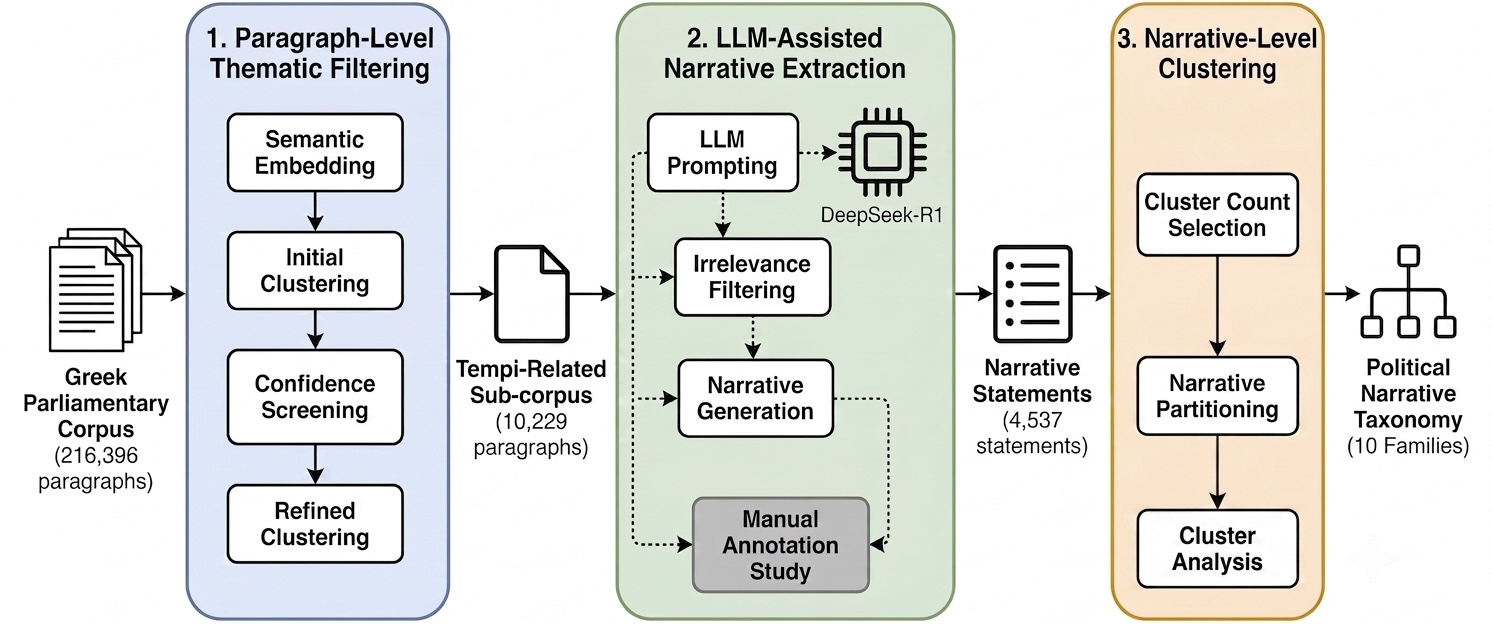}
    \caption{Overview of the \textsc{CART} framework (Pseudocode in Appendix Alg.~\ref{alg:pipeline}).}
    \label{fig:pipeline_workflow}
\end{figure*}

To address this challenge, we propose \textsc{CART} (\textbf{C}lustering-based \textbf{A}nalysis of \textbf{R}ecurring narratives for \textbf{T}axonomy construction), a computational pipeline that combines semantic text representations for paragraph encoding, unsupervised confidence-guided filtering to retain relevant and reliable content, LLM-assisted narrative extraction validated through manual annotation, and narrative clustering followed by cluster analysis to construct a data-driven taxonomy (illustrated in Fig.~\ref{fig:pipeline_workflow}). Our study investigates political narratives surrounding the Tempi train crash through a computational analysis of Greek parliamentary discourse. Specifically, we address two research questions: \textit{(i)} how political narratives can be systematically extracted and organized from large-scale discourse using computational methods, and \textit{(ii)} which narrative families characterize parliamentary debate surrounding the Tempi train crash.

The study makes two contributions. Empirically, it identifies and organizes recurring political narratives surrounding the Tempi train crash into a data-driven taxonomy, providing a structured account of responsibility attribution, institutional failure, railway safety, accountability, privatization, and the political interpretation of the tragedy. Methodologically, it introduces \textsc{CART}, a computational workflow for extracting narrative statements and deriving interpretable narrative families without relying on predefined narrative categories. To facilitate reproducibility, both the implementation and the processed dataset are publicly available at \url{https://github.com/semoglou/cart}.

\section{Related Work}

Most approaches to narrative extraction focus on identifying structural elements such as actors, actions, and events. Traditional NLP methods use tools such as Semantic Role Labeling, Abstract Meaning Representation, and Named Entity Recognition to extract core narrative elements, entity--verb--entity triples, semantic graphs, or specific actors from text~\cite{anantharama2022canarex, jing2021characterizing, pournaki2025extracting, amorim2024text2story}.

More recently, LLM-based approaches have enabled the extraction of character roles, interpretive frames, and broader narrative structures with less domain-specific training data~\cite{otmakhova2025narrative, stammbach2022heroes}. However, they may still rely on predefined categories, carefully designed prompts, or exhaustive processing of large corpora, particularly when only part of the corpus is relevant.

For taxonomy construction, traditional approaches often rely on expert-driven manual coding to identify domain-specific narratives and construct narrative codebooks~\cite{hughes2021development, kotseva2023trend}. Recent work has also developed hierarchical taxonomies that capture both broad narrative categories and finer-grained themes~\cite{coan2021computer, nikolaidis2025polynarrative}. Although interpretable, these approaches can be difficult to scale to large parliamentary corpora and often require predefined categories.

The present study integrates these directions into a unified, largely unsupervised workflow. Existing approaches typically address narrative extraction or taxonomy construction as separate tasks~\cite{anantharama2022canarex, otmakhova2025narrative, coan2021computer, nikolaidis2025polynarrative}. In contrast, our approach combines unsupervised thematic filtering, LLM-assisted extraction of concise narrative statements, and narrative-level clustering to derive narrative families without relying on predefined narrative categories (Fig.~\ref{fig:pipeline_workflow}).

\section{Dataset}\label{subsec:dataset}

The dataset used in this study is based on the Greek Parliament Proceedings, which have been increasingly explored in computational linguistics and political analysis~\cite{barmpounis2024recognition, gavriilidou2024investigating, lislevand2024estimating}. A major resource in this area is the Greek Parliament Proceedings Dataset for Computational Linguistics and Political Analysis by \citet{dritsa2022greek}.\footnote{\url{https://zenodo.org/records/7005201}}
The Greek Parliament was selected because it constitutes a central institutional arena for political discourse: beyond its legislative and representative functions, it also contributes to the shaping of public opinion and political debate~\cite{patz2025analyzing, treimane2011analyzing}.
The dataset spans July 1989 to July 2020 and therefore does not cover the Tempi train crash. To analyze parliamentary discourse after the event, we extended the corpus with more recent Parliament Proceedings, following the data collection and formatting procedures of \citet{dritsa2022greek}.
This ensured consistency with the original dataset and extended the available corpus by five years, from July 2020 to August 2025.

The extended dataset contains 185,857 speech fragments from 907 parliamentary sitting sessions, stored in a CSV file of approximately 534 MB with 11 metadata fields, including speaker name, sitting date, political party, parliamentary session information, government, electoral region, parliamentary role, and gender.
Since the study focuses on the Tempi train crash, we constructed a sub-corpus covering March 7, 2023, the first relevant parliamentary entry after the crash, to August 28, 2025. This sub-corpus contains 88,058 speeches, which were segmented into 216,396 original paragraphs for paragraph-level thematic filtering. Speech ID and paragraph ID fields were added to preserve traceability between paragraphs and source speeches.

\begin{figure*}[t!]
    \centering
    \includegraphics[width=\textwidth]{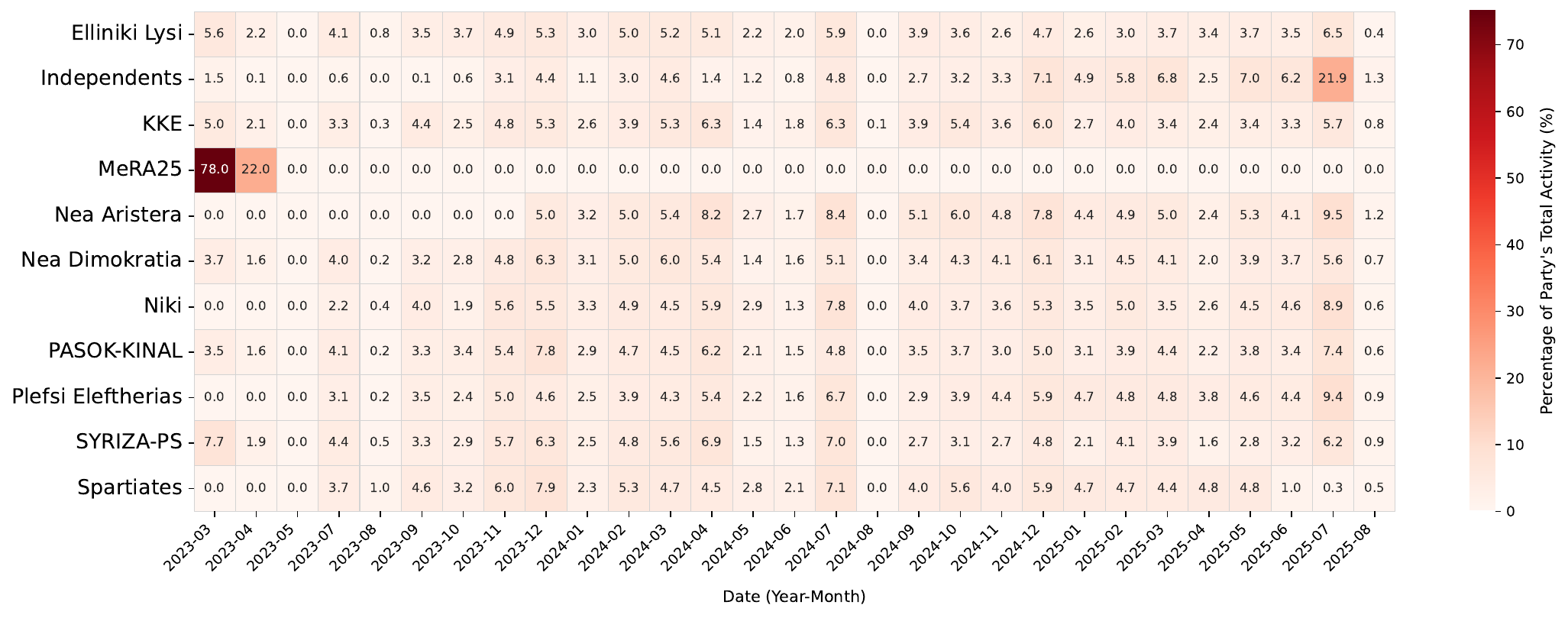}
    \caption{Percentage distribution of paragraph activity across political parties by month.}
    \label{fig:activity_per_month}
\end{figure*}

The paragraph-level corpus was then cleaned by removing procedural or audience-rendered paragraphs, corrupted rows, and paragraphs shorter than 100 characters. Unused metadata fields were dropped, sitting dates were aggregated by month, political party names were standardized in English, and variants of the same party were unified. The cleaned paragraph text was created by lowercasing, punctuation removal, and stress-mark removal. The final analytical dataset contains 117,716 paragraphs with metadata for member name, member gender, sitting date, political party, speech ID, paragraph ID, and cleaned paragraph text. Figure~\ref{fig:activity_per_month} presents monthly paragraph activity across parties; additional political party and gender distribution statistics are provided in Appendix~\ref{app:dataset_distribution}.

\section{Computational Pipeline}

The \textsc{CART} framework consists of three stages that move from parliamentary paragraphs to a narrative taxonomy. First, paragraph-level thematic filtering is used to identify candidate event-related discourse within the larger corpus. Second, LLM-assisted extraction converts relevant paragraphs into concise narrative statements, while excluding paragraphs that do not express a Tempi-specific narrative. Third, narrative-level clustering groups the extracted statements into broader narrative families. This design separates the main analytical units of the study: paragraphs are used for discourse filtering, narrative statements are the product of the extraction, and narrative families form the final taxonomy. Because no predefined narrative categories or relevance labels were available, the pipeline relies primarily on unsupervised methods that exploit the structure of semantic embedding spaces.

\subsection{Paragraph-Level Thematic Filtering}\label{subsec:first}

The first stage aims to reduce the full parliamentary corpus to candidate discourse regions that are likely to contain Tempi-related content. Because no paragraph-level relevance labels were available, we used clustering as an exploratory filtering strategy for grouping semantically related parliamentary paragraphs. Data clustering seeks to identify natural groupings in multidimensional data according to a similarity measure~\cite{omran2007overview}. In text clustering, however, obtaining clusters that are both semantically coherent and geometrically well separated remains challenging, especially in large, heterogeneous corpora~\cite{petukhova2025text}.

Paragraphs were represented using contextual embeddings generated with the Qwen3-Embedding-8B model~\cite{zhang2025qwen3}. Contextual embeddings provide dense semantic representations whose values depend on the surrounding linguistic context, allowing them to capture meaning shifts more effectively than static representations~\cite{jurafsky2026speech}. Since such embeddings are high-dimensional, we applied Principal Component Analysis~\cite{jolliffe2025principal} before clustering; a dimensionality analysis assessing retained variance and nearest-neighbor preservation is provided in Appendix~\ref{app:pca_dimensionality}. 

To group the parliamentary paragraphs into coherent thematic units, we used centroid-based clustering with \textit{k}-means over the PCA-reduced contextual embeddings. This choice provides a scalable and transparent baseline for partitioning a large embedding space. Let $x_i$ denote a paragraph embedding assigned to cluster $C_{c(i)}$. For each paragraph, the Silhouette coefficient~\cite{rousseeuw1987silhouettes} is defined as:
\begin{equation}\label{eq:sil}
s_i=\frac{b_i-a_i}{\max(a_i,b_i)},
\end{equation}
where $a_i$ is the average distance of $x_i$ from the other paragraphs in its assigned cluster, and $b_i$ is the minimum average distance of $x_i$ from the paragraphs of any other cluster. The coefficient captures the balance between \emph{cohesion} and \emph{separation}: values close to $1$ indicate that a paragraph is well matched to its cluster and well separated from competing clusters, values around $0$ indicate boundary cases, and negative values suggest that the paragraph may be better placed in a different cluster~\cite{dudek2020silhouette}.
Since the exact Silhouette coefficient requires pairwise distance computations between all paragraphs, direct evaluation is computationally demanding for large corpora. We therefore used a centroid-based approximation~\cite{wangsilapprox}, replacing $a_i$ and $b_i$ with distances to cluster centroids:
\begin{equation}\label{eq:approx}
\tilde{a}_i=\|x_i-\mu_{c(i)}\|, \quad
\tilde{b}_i=\min_{r\neq c(i)} \|x_i-\mu_r\|,
\end{equation}
where $\mu_{c(i)}$ is the centroid of the assigned cluster and $\mu_r$ is the centroid of cluster $C_r$. This approximation preserves the intuition of contrasting within-cluster cohesion with separation from the nearest competing cluster, while substantially reducing computational cost.

To determine the number of clusters, we evaluated candidate solutions for $k \in \{15,\dots,100\}$. This range was specified in advance as an exploratory search interval, broad enough to allow both general parliamentary themes and more specific event-related subthemes to emerge, while avoiding very small values of $k$ that would merge distinct topics and very large values that would produce highly fragmented clusters. For each candidate value, we monitored two complementary aggregate versions of the Silhouette score. The first was the \emph{micro-averaged Silhouette score}, computed as:
\begin{equation}\label{eq:mis}
S_{\text{micro}}(k)=\frac{1}{n}\sum_{i=1}^{n}s_i,
\end{equation}
where $n$ is the total number of paragraphs. This aggregation gives equal weight to individual paragraphs and therefore reflects the overall geometric quality of the partition. However, because every paragraph contributes equally, the resulting score is naturally influenced more strongly by large clusters~\cite{dudek2020silhouette}. The second was the \emph{macro-averaged Silhouette}, computed as:
\begin{equation}\label{eq:mas}
S_{\text{macro}}(k)=\frac{1}{k}\sum_{r=1}^{k}\left(\frac{1}{|C_r|}\sum_{x_i\in C_r}s_i\right),
\end{equation}
where $k$ is the number of clusters and $|C_r|$ is the size of cluster $C_r$. Here, Silhouette values are first averaged within each cluster and then across clusters, so that each cluster contributes equally regardless of its size. This makes the criterion more sensitive to smaller and potentially more specialized thematic groups~\cite{revisiting}.

Both aggregation views were important for the present corpus. Parliamentary discourse is highly uneven: some themes are broad and recurrent, while others appear in smaller but substantively meaningful concentrations. Optimizing only the micro-averaged Silhouette would privilege high-volume clusters, whereas optimizing only the macro-averaged Silhouette could overemphasize small or less stable groups~\cite{topc}. For this reason, among candidate solutions with positive aggregate scores, we selected the value of $k$ that maximized the harmonic mean of the two criteria:
\begin{equation}\label{eq:hmmima}
H(k)=\frac{2\,S_{\text{micro}}(k)\,S_{\text{macro}}(k)}
{S_{\text{micro}}(k)+S_{\text{macro}}(k)}.
\end{equation}
The harmonic mean was used as a conservative combination rule because it penalizes candidate solutions when either the micro- or macro-averaged score is low, favoring partitions that perform reasonably well under both perspectives. As an internal validation criterion, this score is not treated as ground-truth evidence of topical relevance; rather, it is used to select a geometrically coherent candidate partition for subsequent relevance screening and narrative extraction. Using this criterion, the initial paragraph-level partition was set to $k=52$ (the corresponding selection curve is provided in Appendix~\ref{app:cluster_summaries}; Fig.~\ref{fig:k-selection-strategy}).

After obtaining the initial thematic partition, we applied a confidence diagnostic for paragraph cluster assignments, implemented using the CAKE Python package.\footnote{\url{https://pypi.org/project/cake-ensemble/}.} Additional details on the confidence-scoring procedure are provided in Appendix~\ref{app:cake_distribution}. Paragraphs in the lower quartile of the confidence score distribution were removed. This threshold was used as a conservative heuristic to exclude the least stable assignments while retaining most of the corpus for refinement. The corresponding score distribution is reported in Appendix~\ref{app:cake_distribution}, Fig.~\ref{fig:cake_distribution}. This confidence-based filtering reduced the influence of ambiguous paragraph assignments before the refined thematic partition was constructed.

Using this confidence-filtered subset, we repeated the same cluster-count selection procedure, again using the harmonic mean of the micro- and macro-averaged Silhouette scores. This second pass selected $k=41$. The refined paragraph-level thematic partition was then produced using \textit{k}-means++ initialization to reduce sensitivity to the initial centroid selection.

\subsection{Narrative Extraction}\label{subsec:narext}

Before extracting narrative statements, it was first necessary to identify which of the refined paragraph-level clusters were substantively related to the Tempi train crash. To support this screening step, we constructed cluster-level metadata for each of the $41$ refined clusters; the metadata fields are listed in Table~\ref{tab:cluster_metadata} in Appendix~\ref{app:cluster_metadata}. Specifically, for each cluster, we recorded its total paragraph count, retrieved the five paragraphs with the highest CAKE scores, and extracted the top 20 TF-IDF terms by concatenating all paragraphs assigned to that cluster. The high-score paragraphs served as representative cluster examples, while the TF-IDF terms provided a compact lexical summary.

These metadata were then provided to the DeepSeek-R1 model through the Amazon Bedrock API. The model received the representative paragraphs together with the top TF-IDF terms and was instructed to generate a concise descriptive title for each cluster while identifying clusters directly or plausibly related to the Tempi train crash. In this step, the LLM was used as an interpretive aid for cluster screening, rather than as the source of the final narrative taxonomy.\footnote{Prompts/code: \url{https://github.com/semoglou/cart}.}
Five clusters were identified as directly related to the Tempi train crash or as broader themes likely to contain Tempi-related paragraphs. The LLM-generated titles were:
\textit{Railway Safety and Infrastructure Reform}, \textit{Government Accountability and Distrust}, \textit{Cover-up and Accountability Investigations}, \textit{Tempi Train Crash Accountability Investigation}, \textit{Parliamentary Investigative Committee Formation}.
Together, these clusters contained 10,229 paragraphs, forming the sub-dataset for narrative analysis. The top five TF-IDF terms in each relevant cluster are presented in Appendix~\ref{app:tempi_terms}, Fig.~\ref{fig:tempi_wordcloud}. 

To ensure that the final taxonomy was built from topically relevant material, narrative statement extraction was restricted to this Tempi-focused subset. The same LLM was then used for paragraph-level extraction. For each isolated paragraph, the model was instructed either to generate a single concise narrative statement or to mark the paragraph as irrelevant if it was not substantively focused on the Tempi train crash.\footnote{Narratives were generated directly in English from Greek source paragraphs.} In addition to the paragraph text, the model was provided with the narrative definition established in \S\ref{sec:intro}, together with the sitting date and the party of the speaker.

Ultimately, 4,537 narrative statements were generated, accounting for 44.35\% of the 10,229 paragraphs processed by the model, while the remaining 5,692 paragraphs (55.65\%) were classified as irrelevant to the crash. The narrative yield varied substantially across the five Tempi-related clusters: the two most focused clusters produced narrative statements for 95.25\% and 77.63\% of their paragraphs, while a third cluster yielded narrative statements for 66.67\% of its paragraphs. The remaining two clusters were broader thematic clusters with lower narrative yields, producing narrative statements for 18.26\% and 7.63\% of their paragraphs. This variation indicates that the selected clusters differed in their degree of direct relevance to the Tempi case, justifying the additional paragraph-level relevance filtering performed during extraction. 

To assess the LLM-assisted extraction step, we conducted a manual annotation study on a sample of retained paragraphs and generated narrative statements. Across five annotators, the extracted narrative statement was selected as the best-fitting option in 84.6\% of annotations, with an annotator-level standard error of 1.21 percentage points. Details are provided in Appendix~\ref{app:annotation_study}.\footnote{Annotation materials are available at \url{https://github.com/semoglou/cart}.}

\subsection{Narrative Clustering}\label{sec:narclust}

After the extraction stage, the resulting narrative statements were encoded as embeddings, following the same representation procedure as in the paragraph-level stage, and grouped into broader narrative families to construct the taxonomy. At this stage, each point in the embedding space represented a semantically condensed narrative statement rather than a raw parliamentary paragraph. This made the narrative-level space smaller, less heterogeneous, and more suitable for finer-grained unsupervised analysis.

This difference motivated the use of different clustering procedures. The paragraph-level stage was designed as a large-scale exploratory filtering step, where scalability, transparency, and robustness were prioritized. For that reason, standard \textit{k}-means was used to identify broad thematic regions in the parliamentary corpus. In contrast, the narrative-level stage served as the basis for the final taxonomy, where preserving finer distinctions between semantically related narrative groups was more important. A more specialized silhouette-driven procedure was therefore used at this stage.

To determine the appropriate number of narrative clusters, we employed the Composite Silhouette method.\footnote{\url{https://pypi.org/project/compsil/}.} Additional details on the cluster-selection procedure are provided in Appendix~\ref{app:composite_silhouette}.
This validation procedure was computationally feasible because the extracted narrative corpus was substantially smaller than the paragraph corpus used during thematic filtering, making it practical to assess the stability of candidate solutions across repeated data perturbations. Using this criterion, the selected number of clusters was set to $k=10$.

Having fixed the number of clusters, we partitioned the narrative embeddings using a silhouette-driven variant of \textit{k}-means designed to produce more reliable centroid updates in semantically fine-grained representation spaces, \textsc{K-Sil}~\cite{semoglouksil}. Additional details are provided in Appendix~\ref{app:ksil}.
Under identical experimental conditions, standard \textit{k}-means produced an overall Silhouette score of approximately $0.16$, whereas the \textsc{K-Sil} partition reached approximately $0.20$. This indicates improved cohesion and separation, and we therefore retained the \textsc{K-Sil} solution at $k=10$ as the final partition used to construct the narrative taxonomy (Table~\ref{tab:final_taxonomy}). Cluster summaries for both solutions are provided in Appendix~\ref{app:cluster_summary_comparison}, including prevalence and mean cluster-level Silhouette.

\section{Resulting Narrative Taxonomy}\label{subsec:tax}
The final taxonomy was constructed from the \textsc{K-Sil} partition of the 4,537 extracted narrative statements (Appendix~\ref{app:cluster_summary_comparison}, Fig.~\ref{fig:ksil_cluster_summary}). Each narrative cluster was interpreted using high-silhouette narrative statements together with contrastive terms derived from comparisons with the nearest neighboring clusters. This resulted in ten narrative families, presented in Table~\ref{tab:final_taxonomy}. The taxonomy organizes recurring parliamentary interpretations concerning the crash itself, accountability, investigation procedures, institutional responsibility, railway safety, privatization, and political exploitation. The cluster labels summarize claims expressed in parliamentary discourse rather than making factual assertions about the events. Figure~\ref{fig:temporal_narratives} illustrates the temporal evolution of the discovered narrative families (see Appendix Figs.~\ref{fig:party_temporal_evolution} and~\ref{fig:top1_narrative_by_party} for complementary party-level analyses).

\begin{table*}[t]
\centering
\small
\begin{tabular}{p{0.13\textwidth} p{0.34\textwidth} p{0.45\textwidth}}
\toprule
\textbf{Cluster} & \textbf{Narrative family} & \textbf{Representative narrative} \\
\midrule
\textbf{0} (13.82\%) & \textit{Judicial and procedural obstacles to accountability}
& Forming investigative committees obstructs judicial accountability processes. \\
\midrule
\textbf{1} (23.32\%) & \textit{Government cover-up of Tempi accountability}
& Government-led cover-up obstructs Tempi crash accountability. \\
\midrule
\textbf{2} (13.29\%) & \textit{Systemic governance failures caused preventable disaster}
& Systemic governance failures caused preventable disaster. \\
\midrule
\textbf{3} (5.20\%) & \textit{Tempi crash resulted from systemic criminal failure}
& Tempi crash constitutes criminal systemic failure. \\
\midrule
\textbf{4} (8.42\%) & \textit{Government obstructs crash investigation to evade accountability}
& Government obstructs crash investigation to evade accountability. \\
\midrule
\textbf{5} (7.98\%) & \textit{Negligence caused the preventable Tempi train crash}
& Infrastructure negligence caused preventable Tempi train crash. \\
\midrule
\textbf{6} (18.45\%) & \textit{Railway safety remains neglected and unsecured}
& Railway safety remains unsecured due to governmental failure. \\
\midrule
\textbf{7} (3.13\%) & \textit{Legal framework shields ministers from accountability}
& Legal framework shields ministers from Tempi crash accountability. \\
\midrule
\textbf{8} (3.75\%) & \textit{Opposition exploits the tragedy for political gain}
& Opposition exploits tragedy for political gain. \\
\midrule
\textbf{9} (2.64\%) & \textit{Railway privatization caused systemic safety failures}
& Railway privatization caused systemic safety failures. \\
\bottomrule
\end{tabular}
\caption{Final narrative taxonomy produced from the \textsc{K-Sil} clustering solution (Appendix Fig.~\ref{fig:ksil_cluster_summary}). The narrative-family labels paraphrase recurring claims expressed by parliamentary speakers and should not be interpreted as factual claims by the authors. Percentages in parentheses indicate each cluster's share of the extracted narrative corpus, while the final column reports the highest-silhouette representative statement for each cluster. Additional representative narratives for each cluster are provided in Appendix Fig.~\ref{fig:distinctive_representatives}.}
\label{tab:final_taxonomy}
\end{table*}

\begin{figure*}[t]
    \centering
    \includegraphics[width=\textwidth]{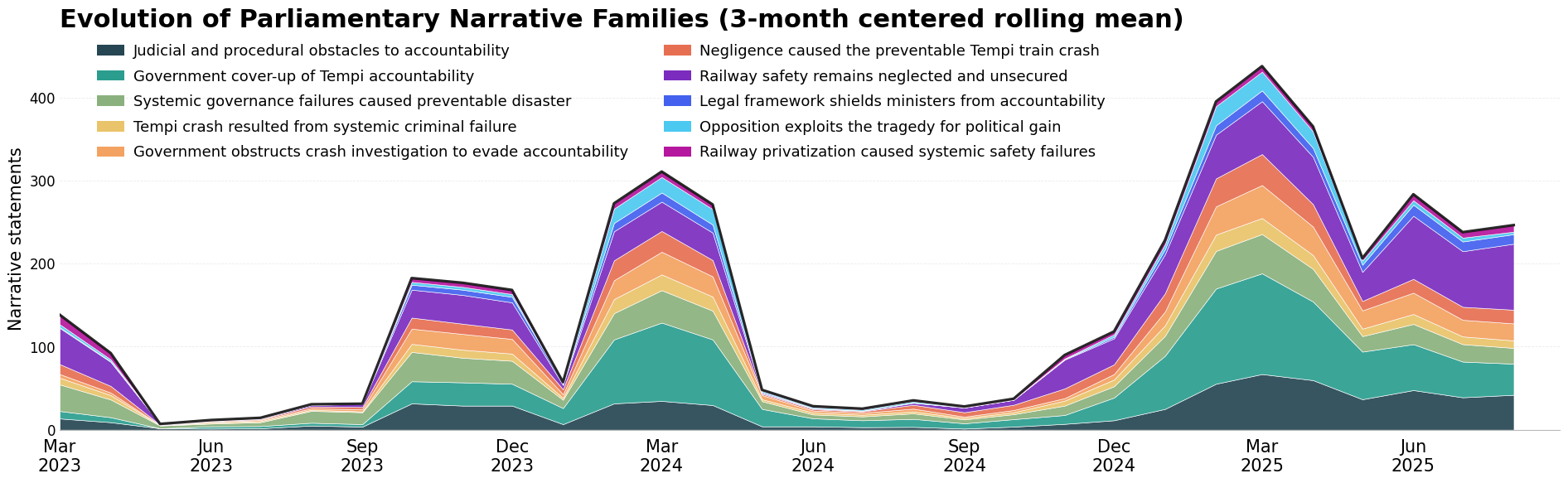}
    \caption{
        \textbf{Temporal distribution of parliamentary narrative families (3-month centered rolling mean).}
        The stacked area chart shows the monthly prevalence of the
        narrative clusters across the parliamentary corpus.
        Each colored band corresponds to one narrative family, and its vertical
        thickness at a given point indicates the number of narrative statements assigned
        to that family.
    }
    \label{fig:temporal_narratives}
\end{figure*}

The largest narrative family is Cluster 1, \emph{Government cover-up of Tempi accountability}, accounting for 23.32\% of the extracted narrative corpus. It frames the investigation as being obstructed through concealment, institutional manipulation, or interference with accountability processes, expressing a government-led cover-up interpretation of the Tempi crash. The second-largest family is Cluster 6, \emph{Railway safety remains neglected and unsecured} (18.45\%). It shifts attention from the investigation to the condition of the railway system, emphasizing neglected safety mechanisms, non-operational infrastructure, and insufficient reform.

Clusters 0, 2, and 5 form a broader group of accountability and causality narratives. Cluster 0, \emph{Judicial and procedural obstacles to accountability}, focuses on the institutional and legal processes through which responsibility is investigated, emphasizing judicial evidence, investigative committees, criminal accountability, and procedural constraints. Cluster 2, \emph{Systemic governance failures caused preventable disaster}, presents the crash as the outcome of broader governance failures rather than an isolated event. Cluster 5, \emph{Negligence caused the preventable Tempi train crash}, links infrastructure negligence, governmental incompetence, and preventability directly to the crash.

\begin{figure}[h]
    \centering
    \includegraphics[width=\columnwidth]{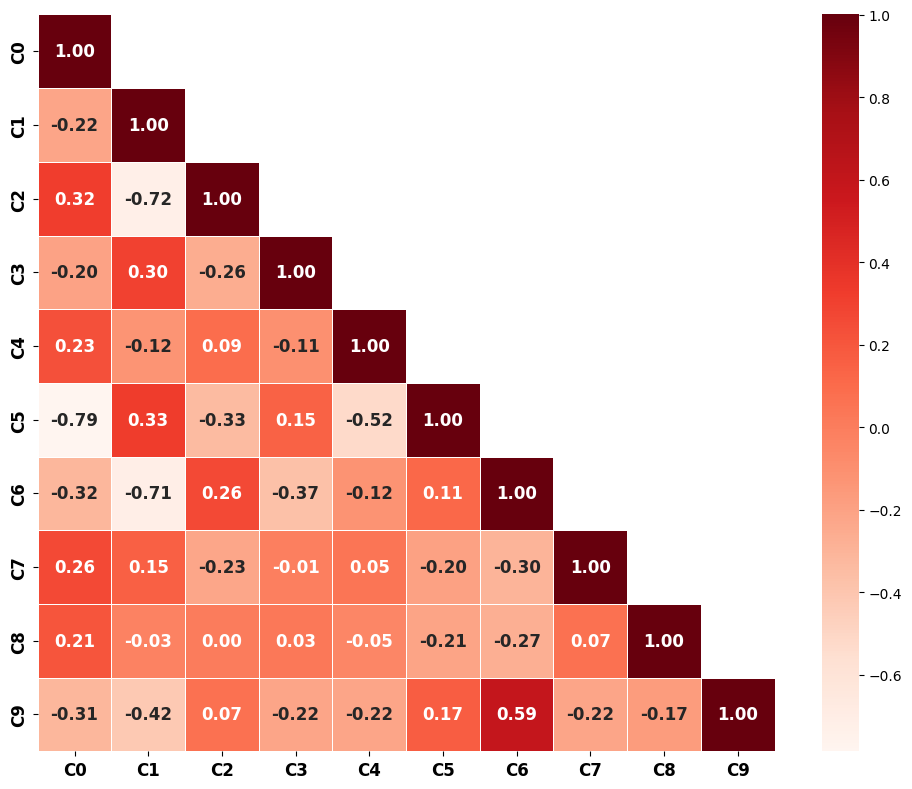}
    \caption{Centroid cosine similarity matrix for the ten
    narrative families. Higher values indicate greater
    semantic proximity between cluster centroids.}
    \label{fig:centroid_sim}
\end{figure}

A second set of clusters captures more specific responsibility-oriented narratives. Cluster 3, \emph{Tempi crash resulted from systemic criminal failure}, frames the disaster as the result of unresolved systemic and potentially criminal failures. Cluster 7, \emph{Legal framework shields ministers from accountability}, focuses on ministerial responsibility and legal or constitutional mechanisms presented as limiting criminal liability. Cluster 9, \emph{Railway privatization caused systemic safety failures}, attributes railway safety failures to privatization, fragmentation, and insufficient investment guarantees.
Cluster 8, \emph{Opposition exploits the tragedy for political gain}, differs from the other families by framing the tragedy as politically instrumentalized by opposition parties rather than assigning responsibility for the crash or its investigation. It also has the highest mean Silhouette score, making it the most geometrically distinct narrative family.
Because several clusters were semantically close, we examined their semantic proximity using centroid cosine similarity (Fig.~\ref{fig:centroid_sim}), complemented by contrastive TF-IDF terms to distinguish closely related narrative families. The closest pair was Clusters 6 and 9 (cosine similarity 0.590). Although both concern railway safety failures, Cluster 6 emphasizes ongoing safety neglect, whereas Cluster 9 attributes these failures specifically to privatization and railway fragmentation. Likewise, Cluster 1 captures a broader cover-up narrative, while Cluster 4 focuses on obstruction of the investigation. Cluster 2 expresses a systemic-governance interpretation, whereas Cluster 5 directly attributes the Tempi crash to negligence.

The temporal distribution in Fig.~\ref{fig:temporal_narratives} suggests that the salience of the narrative families changed over the course of the parliamentary debate. Activity in March 2023 appears to reflect the immediate framing of the crash, with responsibility, preventability, and railway safety already prominent. The renewed increase in late 2023 coincides with greater visibility of institutional and procedural narratives concerning the investigation, accountability mechanisms, and legal responsibility. 
A stronger peak around the first anniversary in March 2024 indicates renewed parliamentary attention to accountability and railway safety. The largest increase in early 2025 reflects a broader intensification of the debate: governmental responsibility and railway safety remain prominent, while ministerial accountability, privatization, and political exploitation become more visible. 
Overall, the pattern is consistent with a shift from immediate interpretations of the crash toward a debate over institutional responsibility, legal accountability, and its broader political meaning.

Taken together, the results show that parliamentary discourse surrounding the Tempi train crash is organized around multiple recurring narrative families that capture distinct yet overlapping interpretations of responsibility, institutional failure, accountability, and political meaning.

\section{Conclusion}

This study investigated recurring political narratives surrounding the Tempi train crash through a computational analysis of Greek Parliament proceedings. To this end, we introduced \textsc{CART}, a computational pipeline for extracting narrative statements and organizing them into a data-driven taxonomy without relying on predefined narrative categories.
The taxonomy identified ten narrative families spanning accountability, safety, systemic governance failures, negligence, privatization, and the political interpretation of the tragedy. Their prominence also evolved over time, reflecting changes in parliamentary debate. Together, these findings provide a structured account of the recurring political interpretations surrounding the Tempi crash and illustrate how \textsc{CART} supports their systematic extraction and organization.
Methodologically, the analysis highlights the complementary roles of confidence-informed filtering, balanced cluster selection, and silhouette-driven clustering in constructing the resulting narrative taxonomy. 

More broadly, \textsc{CART} is designed for text collections containing recurring interpretations of a target issue. Its application requires appropriate text preprocessing and sufficient recurring interpretive content, while contextual metadata can support extraction and subsequent analyses where available. By deriving narrative families directly from textual data rather than predefined categories, \textsc{CART} provides a flexible framework for systematically identifying and organizing recurring narratives across diverse textual corpora.

\section{Limitations}

As with other unsupervised approaches, the outcomes of \textsc{CART} depend partly on the semantic representations and clustering configurations employed. The framework addresses these methodological considerations through systematic cluster selection, confidence-informed refinement, and clustering diagnostics. Nevertheless, alternative embedding models or parameter configurations may yield different levels of narrative granularity.

The LLM-assisted extraction stage also involves modeling choices, particularly regarding the language model and prompting strategy. Although the extracted narrative statements were assessed through manual annotation, further comparisons across model configurations could provide additional evidence of their consistency.

\section*{Acknowledgments}
This work has been partially supported by project MIS 5154714 of the National Recovery and Resilience Plan Greece 2.0 funded by the European Union under the NextGenerationEU Program.

\paragraph{Author Contributions.} \textbf{AS}: Methodology; Formal Analysis; Software; Writing (Original Draft); Validation; Visualization.  
\textbf{TT}: Methodology; Software; Writing (Original Draft); Investigation. 
\textbf{JP}: Conceptualization; Methodology; Writing (Review/Editing); Supervision.

\bibliography{custom}

\appendix

\section{Dataset Statistics}
\label{app:dataset_distribution}

The corpus is unevenly distributed across political parties. Nea Dimokratia accounts for the largest share of paragraphs (36.87\%), followed by SYRIZA-PS (12.04\%), PASOK-KINAL (11.63\%), KKE (8.08\%), Plefsi Eleftherias (7.94\%), and Elliniki Lysi (7.37\%). The remaining political parties have smaller shares: Nea Aristera (5.16\%), Niki (4.63\%), Spartiates (3.52\%), Independents (2.44\%), and MeRA25 (0.31\%). The corpus is also gender-imbalanced, with male speakers accounting for 77.43\% of the paragraphs and female speakers for 22.57\%.

\section{Cluster Metadata Columns}
\label{app:cluster_metadata}

Table~\ref{tab:cluster_metadata} lists the cluster-level metadata fields used during the Tempi-related cluster screening step described in Section~\ref{subsec:narext}. These fields were constructed for each of the $41$ refined paragraph-level clusters and were provided to the LLM to support cluster titling and relevance screening before narrative statement extraction.

\begin{table}[h] 
\centering
\footnotesize
\begin{tabular}{p{0.34\linewidth} p{0.54\linewidth}}
\toprule
\textbf{Column} & \textbf{Description} \\
\midrule
\texttt{Cluster\_ID} & Index of each cluster. \\
\texttt{Cluster\_title} & LLM-generated title based on the representative paragraphs and the TF-IDF terms. \\
\texttt{Doc\_Count} & Number of paragraphs in each cluster. \\
\texttt{Top\_TFIDF\_terms} & Top 20 TF-IDF terms of each cluster. \\
\texttt{Top\_Paragraphs} & Top 5 paragraphs of each cluster based on CAKE score. \\
\texttt{Top\_CAKE\_Scores} & CAKE scores of the top paragraphs. \\
\bottomrule
\end{tabular}
\caption{Description of cluster-level metadata columns}\label{tab:cluster_metadata}
\end{table}

\section{Cluster Count Selection}
\label{app:cluster_summaries}

\begin{figure}[h]
    \centering
    \includegraphics[width=0.49\textwidth]{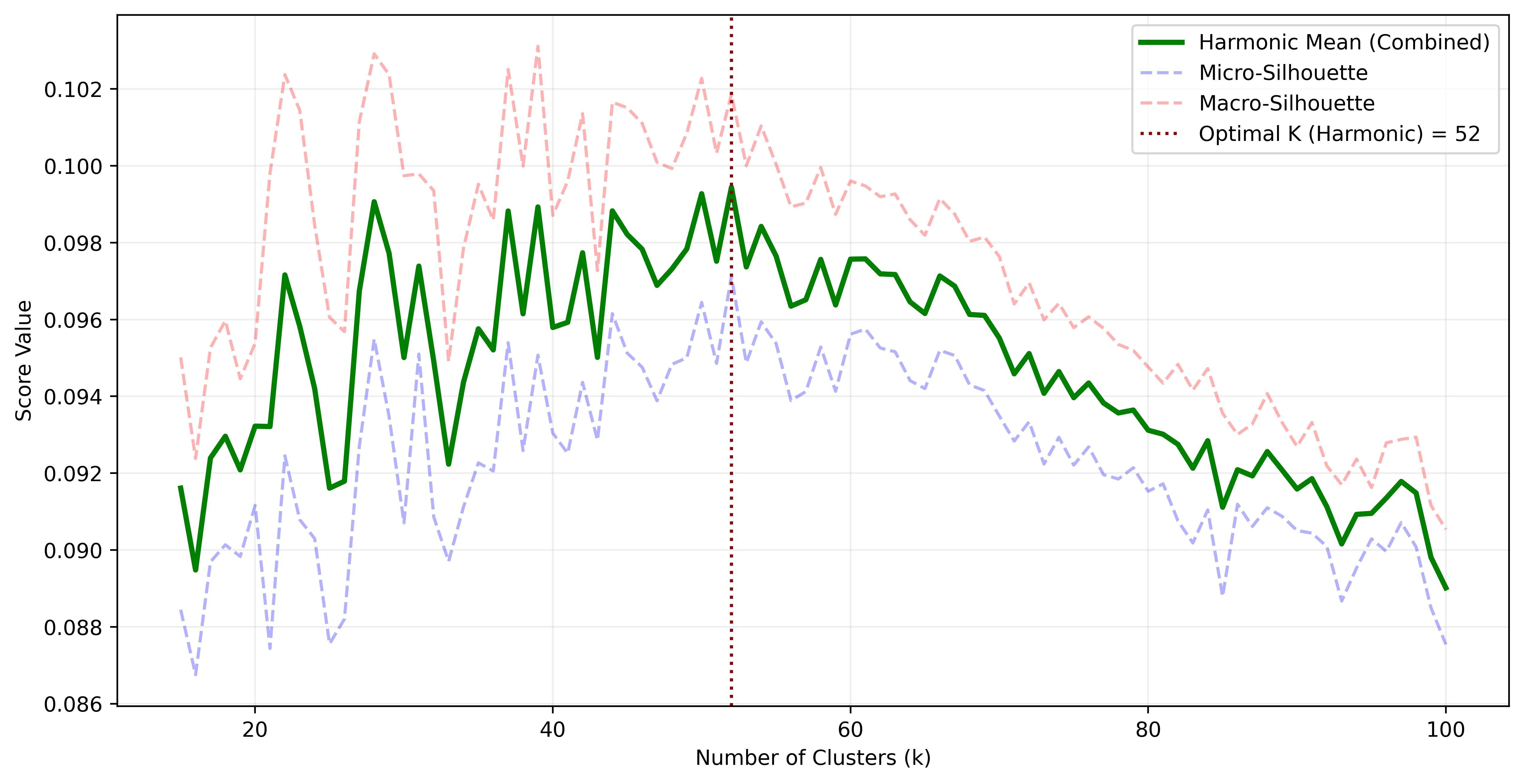}
    \caption{Selection of the number of clusters $k$ using the \textcolor{ForestGreen}{harmonic mean $H(k)$} (Eq.~\ref{eq:hmmima}) of \textcolor{blue}{$S_{\text{micro}}(k)$} (Eq.~\ref{eq:mis}) and \textcolor{red}{$S_{\text{macro}}(k)$} (Eq.~\ref{eq:mas}).}
    \label{fig:k-selection-strategy}
\end{figure}

Figure~\ref{fig:k-selection-strategy} reports the cluster-count selection curve used for the initial paragraph-level thematic filtering stage described in Section~\ref{subsec:first}. Candidate partitions were evaluated over $k \in \{15,\dots,100\}$ using the harmonic-mean criterion $H(k)$ defined in Eq.~\ref{eq:hmmima}, which combines the micro- and macro-averaged Silhouette scores (Eqs.~\ref{eq:mis}--\ref{eq:mas}). This criterion selected $k=52$ for the initial thematic partition before confidence-based refinement.

\section{CAKE Confidence Scoring}
\label{app:cake_distribution}

\begin{figure*}[t]
    \centering
    \includegraphics[width=\textwidth]{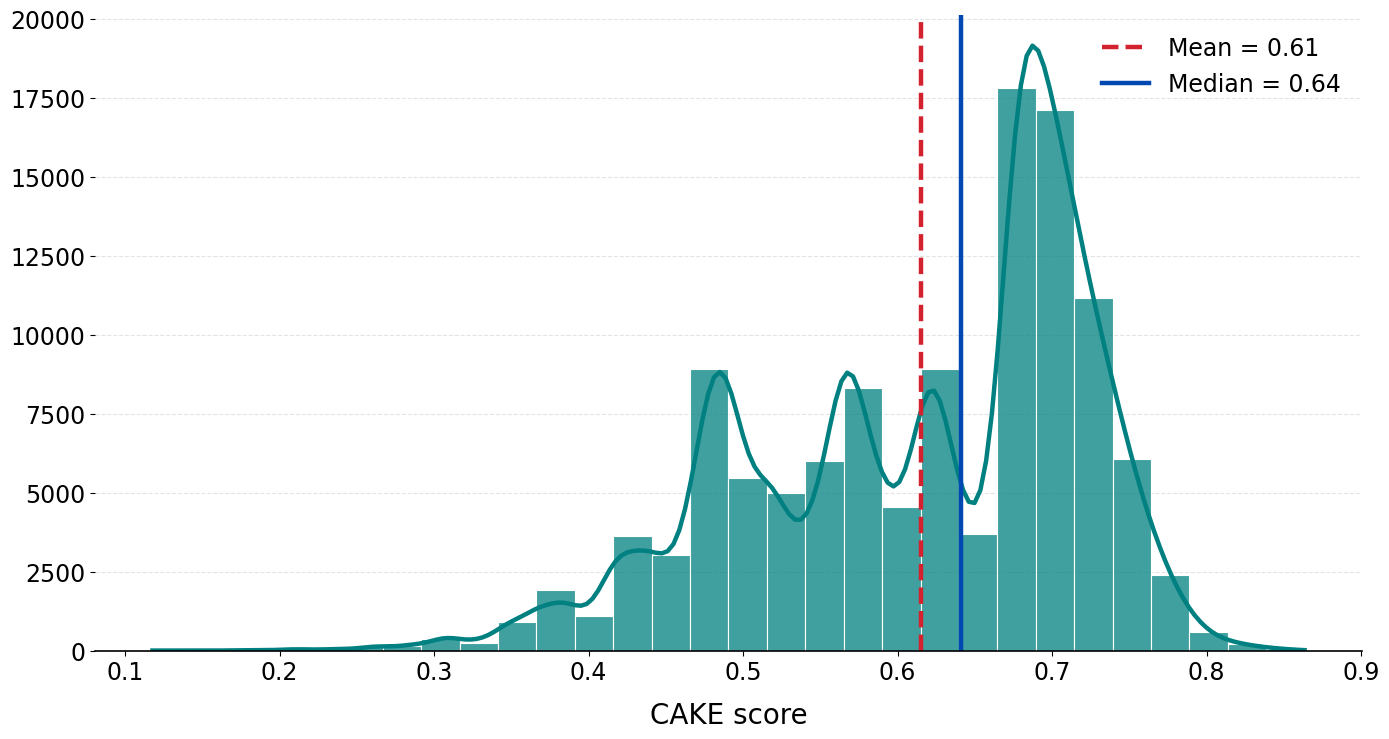}
    \caption{Distribution of CAKE confidence scores across parliamentary paragraphs.}
    \label{fig:cake_distribution}
\end{figure*}

To assess the reliability of paragraph-level cluster assignments in Section \ref{subsec:first}, we employed CAKE~\cite{semogloucake}, implemented using the CAKE package.\footnote{\url{https://pypi.org/project/cake-ensemble/}.} The method jointly estimates how consistently each paragraph is assigned and how well it is supported by the learned clustering structure across repeated clustering runs. In the present study, confidence scores were computed from an ensemble of 30 \textit{k}-means runs.
The resulting confidence scores were used as a post-clustering filtering criterion. Paragraphs in the lower quartile of the confidence score distribution were removed before constructing the refined thematic partition.
Figure~\ref{fig:cake_distribution} reports the distribution of CAKE confidence scores obtained for the parliamentary paragraphs during the initial thematic clustering stage. The distribution illustrates the confidence values used to identify uncertain paragraph assignments prior to the refinement step.

\section{PCA Dimensionality Analysis}
\label{app:pca_dimensionality}

Table~\ref{tab:pca_dimensionality} reports a PCA dimensionality analysis for the paragraph embeddings used in the paragraph-level thematic filtering stage described in Section~\ref{subsec:first}. Rather than evaluating PCA as a clustering-score optimization step, this analysis evaluates it as a representation-preservation step. We report the proportion of retained variance and nearest-neighbor preservation relative to the original embedding space, computed as Recall@10 on a random subsample of 10,000 paragraphs. We retained PCA-200 in the main analysis as a conservative intermediate representation: it preserves 71.60\% of the embedding variance and 81.08\% of original-space nearest neighbors, while avoiding the additional cost of PCA-300 and PCA-400.

\begin{table}[h]
\centering
\small
\begin{tabular}{lccc}
\toprule
\textbf{Rep.} & \textbf{Var.} & \textbf{NN R@10} & \textbf{Time (s)} \\
\midrule
PCA-25  & 0.3117 & 0.3438 & 5.70 \\
PCA-50  & 0.4337 & 0.5166 & 5.10 \\
PCA-100 & 0.5726 & 0.6741 & 9.25 \\
PCA-150 & 0.6574 & 0.7588 & 8.02 \\
PCA-200 & 0.7160 & 0.8108 & 11.31 \\
PCA-250 & 0.7592 & 0.8452 & 13.14 \\
PCA-300 & 0.7924 & 0.8719 & 15.46 \\
PCA-400 & 0.8403 & 0.9071 & 18.57 \\
\bottomrule
\end{tabular}
\caption{PCA dimensionality analysis for paragraph embeddings. NN Recall@10 measures nearest-neighbor preservation relative to the original embedding space on a random subsample of 10,000 paragraphs.}
\label{tab:pca_dimensionality}
\end{table}

\section{Paragraph-Cluster Terms}
\label{app:tempi_terms}

\begin{figure*}[t]
\centering
\includegraphics[width=\textwidth]{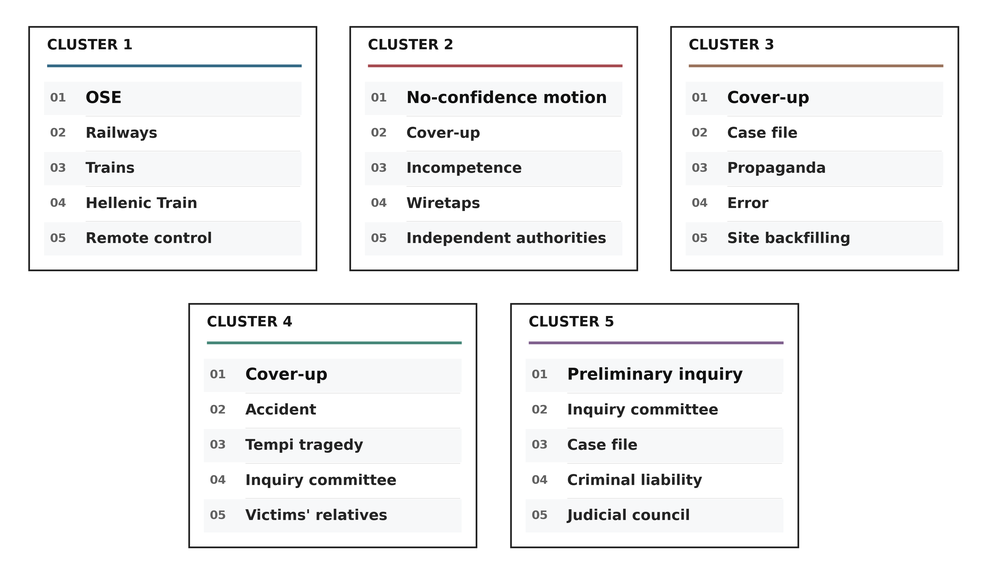}
\caption{Clustered term map of the five Tempi-related parliamentary themes. Terms were selected from the highest-ranking Tempi-relevant TF-IDF features of each cluster and translated into English for readability. Color grouping indicates cluster membership, while larger terms correspond to higher relative TF-IDF relevance within their respective cluster.}
\label{fig:tempi_wordcloud}
\end{figure*}

Figure~\ref{fig:tempi_wordcloud} presents the top TF-IDF terms for the five paragraph-level clusters identified as directly or plausibly related to the Tempi train crash during the narrative extraction stage described in Section~\ref{subsec:narext}. These clusters were produced by the refined paragraph-level thematic partition before narrative extraction. The figure therefore summarizes the lexical content of the paragraph clusters used to select the Tempi-focused subset of 10,229 paragraphs.

\section{Annotation Study}
\label{app:annotation_study}

To assess the narrative extraction procedure described in Section~\ref{subsec:narext}, we conducted a forced-choice validation study on a sample of 100 paragraphs. For each paragraph, annotators were shown the original parliamentary paragraph together with four options: the extracted narrative statement, a randomly sampled narrative statement from another paragraph, an intentionally false LLM-generated narrative statement, and a ``None of the above'' option. Annotators were asked to select the option that best captured the narrative expressed in the paragraph.\footnote{Annotation materials and analysis are available at \url{https://github.com/semoglou/cart}.}

Five native Greek-speaking annotators (three with academic backgrounds in the humanities or social sciences and two with computational backgrounds) completed the task, yielding 500 total annotations. Since each annotator evaluated 100 paragraphs, the annotator-level means in Table~\ref{tab:annotation_results} correspond directly to percentages. Overall, the extracted narrative statement was selected in 84.6\% of annotations. At the annotator level, the mean extracted-selection rate was 84.6\%, with a standard deviation of 2.70 percentage points and a standard error of the mean of 1.21 percentage points.

\begin{table}[h]
\centering
\small
\begin{tabular}{lccc}
\toprule
\textbf{Option} & \textbf{Mean (\%)} & \textbf{SD} & \textbf{SEM} \\
\midrule
Extracted narrative & 84.60 & 2.70 & 1.21 \\
Random narrative    & 4.40  & 2.07 & 0.93 \\
False narrative     & 4.80  & 1.30 & 0.58 \\
None of the above   & 6.20  & 1.30 & 0.58 \\
\bottomrule
\end{tabular}
\caption{Annotation results for the narrative validation task. Mean, Standard Deviation (SD), and Standard Error of the Mean (SEM) are computed across the five annotators.}
\label{tab:annotation_results}
\end{table}

The same pattern was observed under majority-vote aggregation over the 100 annotated paragraphs. The extracted narrative was selected as the majority label for 86 paragraphs, while the random, false, and ``None of the above'' options were selected for 4, 4, and 6 paragraphs, respectively.

As a secondary diagnostic, mean pairwise raw agreement among annotators was 89.5\%, and mean pairwise Cohen's $\kappa$ was 0.616. These values are reported only as supplementary evidence, since the primary purpose of the study was to evaluate whether annotators selected the extracted narrative statement over the alternatives.

These results suggest that, under a constrained forced-choice validation design, the extracted narrative statements were usually selected as the best-fitting option. The lower selection rates for the random, false, and ``None of the above'' options indicate that the task retained meaningful distractors and ambiguous cases, while the extracted statements were selected substantially more often than the alternatives.

A supplementary analysis identified annotator disagreement in 26 of the 100 paragraphs. Non-extracted selections included all distractor categories (31 ``None of the above’’, 24 false narratives, and 22 random narratives). Extracted–random narrative TF-IDF similarity was higher in paragraphs with disagreement (0.0293 vs. 0.0119), with a moderate positive association between similarity and disagreement (Spearman’s $\rho=0.306$).

\section{Composite Silhouette Cluster Selection}
\label{app:composite_silhouette}

To determine the number of narrative clusters, as described in Section \ref{sec:narclust}, we employed Composite Silhouette~\cite{semogloucomposil}, implemented using the Composite Silhouette package.\footnote{\url{https://pypi.org/project/compsil/}.} The method extends the balancing strategy adopted during paragraph-level clustering by jointly considering the micro- and macro-averaged Silhouette scores. The former reflects the overall geometric quality of the partition, whereas the latter gives equal weight to smaller clusters that may correspond to less frequent but analytically meaningful narratives.
Rather than combining these complementary perspectives through a fixed harmonic mean, Composite Silhouette evaluates and adaptively combines them across repeated clustering runs on subsampled data, reducing sensitivity to any single sample or aggregation strategy.

In the present study, this validation procedure selected $k=10$ as the final number of clusters.

\section{Silhouette-Weighted Clustering}
\label{app:ksil}

To construct the final narrative taxonomy (\S~\ref{sec:narclust}), we employed \textsc{K-Sil}~\cite{semoglouksil}, a silhouette-driven instance-weighted variant of \textit{k}-means.
Unlike standard \textit{k}-means, which updates cluster centroids as unweighted means of all assigned instances, \textsc{K-Sil} reweights each instance according to a silhouette-based confidence signal. In \textsc{K-Sil}, centroid updates are computed as softmax-weighted means of the assigned instances with an adaptive temperature parameter that controls the sharpness of the weighting distribution. Embeddings that are more clearly associated with their assigned cluster therefore exert greater influence on centroid updates, whereas embeddings located in more ambiguous boundary regions contribute less. This weighting scheme is well suited to narrative data, where some statements express a distinctive interpretation while others lie near the boundary between related themes.

\section{Narrative Cluster Summaries}
\label{app:cluster_summary_comparison}

\begin{figure*}[t]
    \centering
    \includegraphics[width=\linewidth]{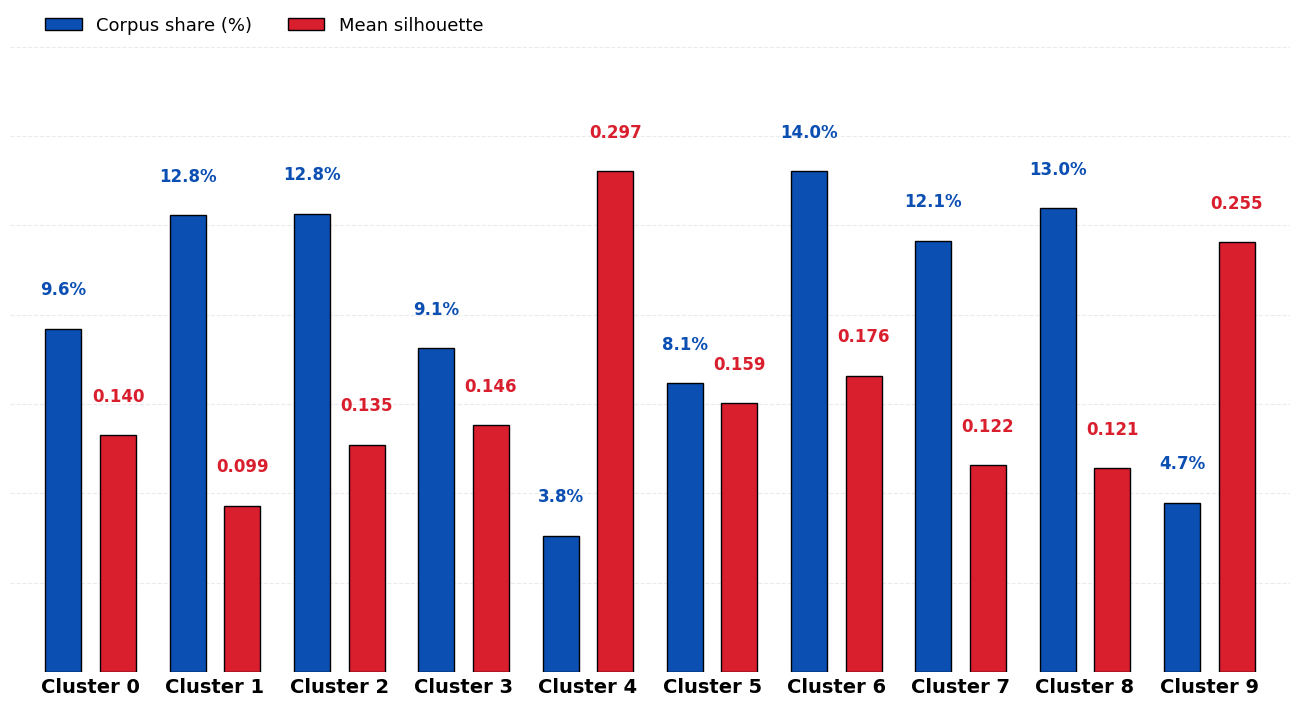}
    \caption{Cluster prevalence and mean cluster-level Silhouette for the ten narrative clusters produced by the standard \textit{k}-means baseline at $k=10$. \textcolor{blue}{Blue bars} show the corpus share (\%) of each cluster, while \textcolor{red}{red bars} show the mean Silhouette score (Eqs.~\ref{eq:sil}, \ref{eq:approx}) of each cluster.}
    \label{fig:kmeans_cluster_summary}
\end{figure*}

\begin{figure*}[t]
    \centering
    \includegraphics[width=\linewidth]{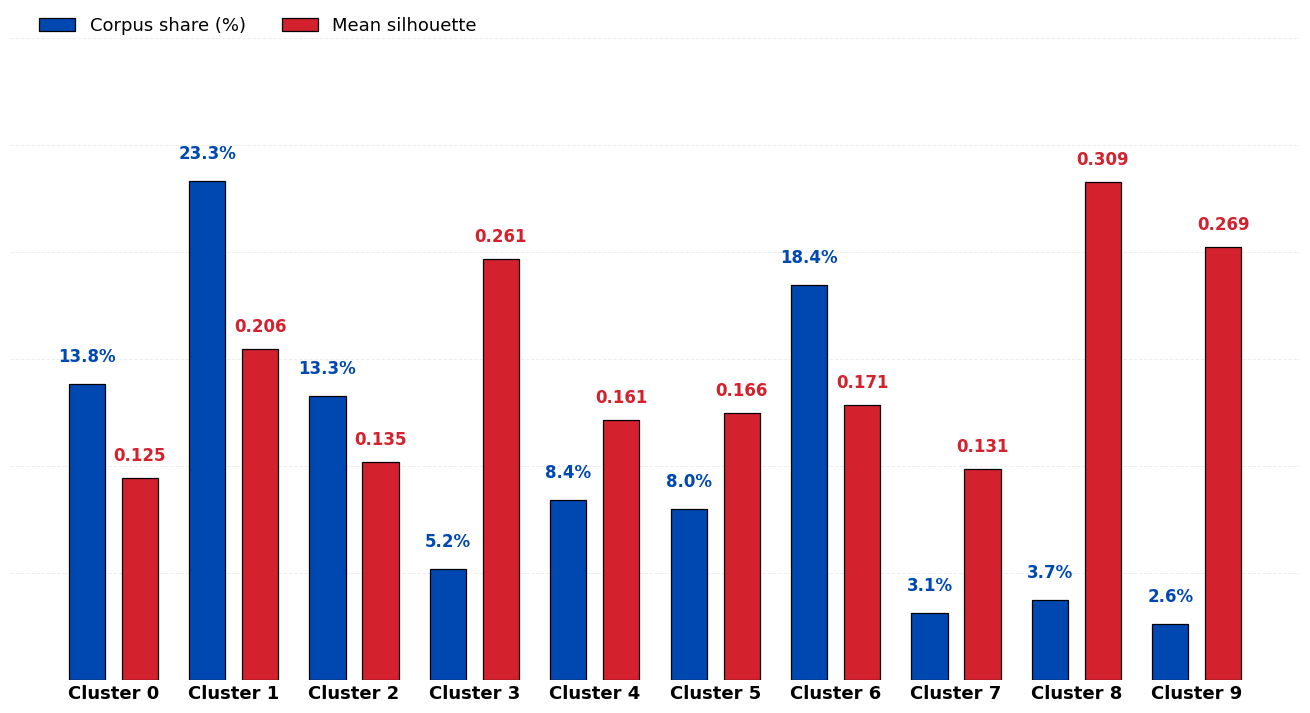}
    \caption{Cluster prevalence and mean cluster-level Silhouette for the ten narrative clusters produced by the \textsc{K-Sil} partition at $k=10$. \textcolor{blue}{Blue bars} show the corpus share (\%) of each cluster, while \textcolor{red}{red bars} show the mean Silhouette score (Eqs.~\ref{eq:sil}, \ref{eq:approx}) of each cluster.}
    \label{fig:ksil_cluster_summary}
\end{figure*}

Figure~\ref{fig:kmeans_cluster_summary} presents the cluster-level summary for the standard \textit{k}-means baseline at $k=10$, while Figure~\ref{fig:ksil_cluster_summary} presents the corresponding summary for the final \textsc{K-Sil} partition. The comparison provides additional context for the aggregate Silhouette improvement reported in the main text, showing differences in both cluster prevalence and mean cluster-level Silhouette.

\begin{figure*}[t]
    \centering
    \includegraphics[width=\textwidth]{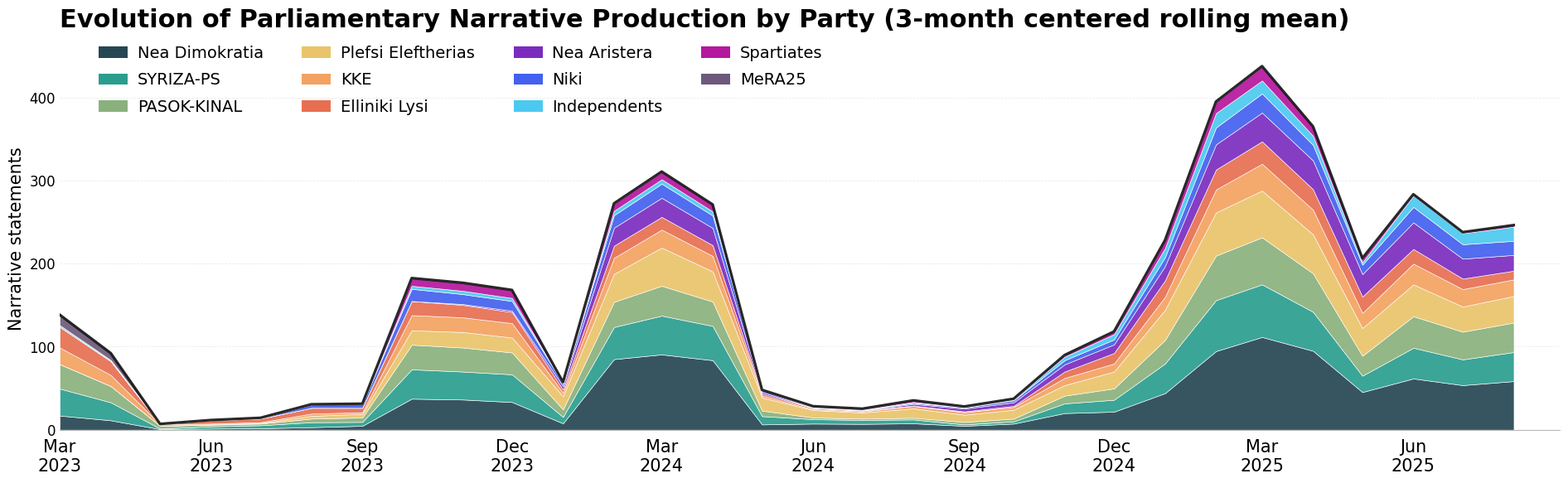}
    \caption{
    Temporal evolution of narrative production by parliamentary party. The stacked areas represent the monthly number of narrative statements attributed to each party, while the black line indicates the total volume of narrative statements across all parties. Values are smoothed using a three-month centered rolling mean to emphasize broader temporal trends.
    }
    \label{fig:party_temporal_evolution}
\end{figure*}

\begin{figure*}[t]
    \centering
    \includegraphics[width=\textwidth]{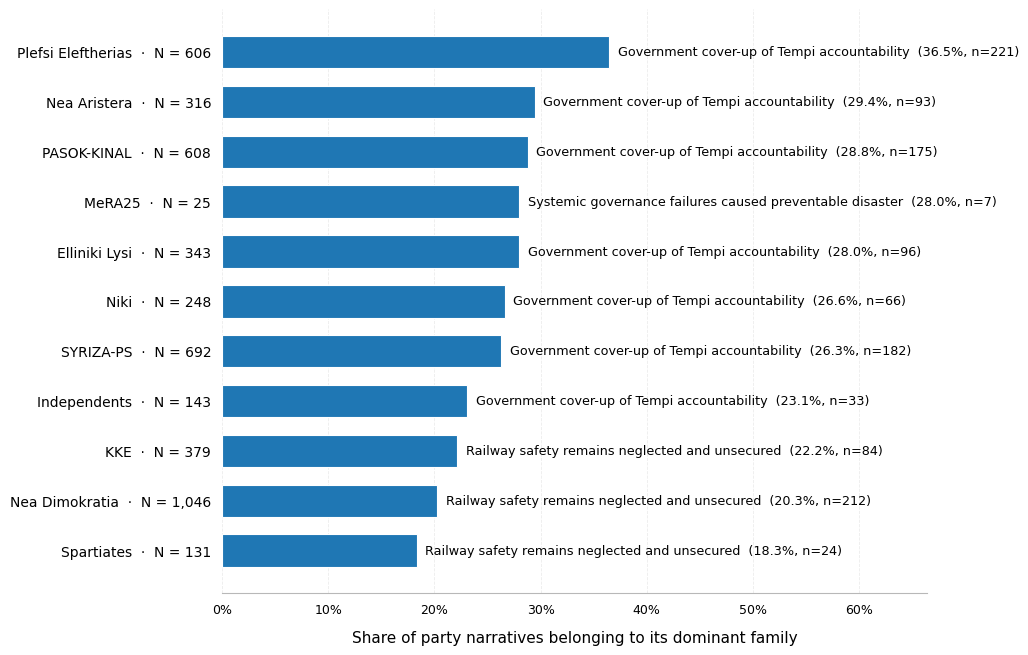}
    \caption{
    Dominant narrative family by parliamentary party. Each horizontal bar represents a parliamentary party, with its length indicating the percentage of the party's narrative statements assigned to its most prevalent narrative family. Labels identify the dominant narrative family together with its percentage share and the corresponding number of narrative statements.
    }
    \label{fig:top1_narrative_by_party}
\end{figure*}

Figure~\ref{fig:party_temporal_evolution} illustrates the temporal evolution of narrative production across parliamentary parties, complementing the narrative-family dynamics presented in the main text by showing how overall narrative activity varied across parties over time. Figure~\ref{fig:top1_narrative_by_party} further summarizes the dominant narrative family for each parliamentary party by reporting the proportion of narrative statements assigned to its most prevalent narrative family, providing an overview of the principal narrative orientation of each party.

\begin{figure*}[t]
\centering
\footnotesize

\begin{minipage}[t]{0.48\textwidth}

\begin{tcolorbox}[narrativebox,title={Judicial and procedural obstacles to accountability}]
\begin{enumerate}[leftmargin=*, itemsep=1pt, topsep=1pt]
    \item Judicial investigation must precede political accountability determinations.
    \item Forming investigative committees obstructs judicial accountability processes.
    \item Government assumes political accountability amid judicial investigation.
\end{enumerate}
\end{tcolorbox}

\begin{tcolorbox}[narrativebox,title={Government cover-up of Tempi accountability}]
\begin{enumerate}[leftmargin=*, itemsep=1pt, topsep=1pt]
    \item Government cover-up obstructs Tempi crash accountability.
    \item Government cover-up obstructs Tempi crash accountability investigation.
    \item Government cover-up obstructed Tempi crash accountability.
\end{enumerate}
\end{tcolorbox}

\begin{tcolorbox}[narrativebox,title={Systemic governance failures caused preventable disaster}]
\begin{enumerate}[leftmargin=*, itemsep=1pt, topsep=1pt]
    \item Systemic governance failures caused preventable disaster.
    \item Systemic governance failures caused preventable disasters via negligence.
    \item Safety failures under government responsibility caused preventable disaster.
\end{enumerate}
\end{tcolorbox}

\begin{tcolorbox}[narrativebox,title={Tempi crash resulted from systemic criminal failure}]
\begin{enumerate}[leftmargin=*, itemsep=1pt, topsep=1pt]
    \item Tempi crash resulted from systemic governance failures.
    \item Tempi crash requires deep investigation to end ambiguities.
    \item Tempi crash resulted from systemic criminal negligence.
\end{enumerate}
\end{tcolorbox}

\begin{tcolorbox}[narrativebox,title={Government obstructs crash investigation to evade accountability}]
\begin{enumerate}[leftmargin=*, itemsep=1pt, topsep=1pt]
    \item Government obstructs crash investigation to evade accountability.
    \item Government obstructs crash investigation to conceal accountability.
    \item Government obstructs crash investigation to avoid accountability.
\end{enumerate}
\end{tcolorbox}

\end{minipage}
\hfill
\begin{minipage}[t]{0.48\textwidth}

\begin{tcolorbox}[narrativebox,title={Negligence caused the preventable Tempi train crash}]
\begin{enumerate}[leftmargin=*, itemsep=1pt, topsep=1pt]
    \item Systemic negligence caused preventable Tempi train crash.
    \item Infrastructure negligence caused preventable Tempi train crash.
    \item Government negligence caused preventable Tempi train crash.
\end{enumerate}
\end{tcolorbox}

\begin{tcolorbox}[narrativebox,title={Railway safety remains neglected and unsecured}]
\begin{enumerate}[leftmargin=*, itemsep=1pt, topsep=1pt]
    \item Systemic railway safety failures are widespread across EU.
    \item Railway safety not contingent on specific system's existence.
    \item Railway safety failures pose lethal risks to passengers.
\end{enumerate}
\end{tcolorbox}

\begin{tcolorbox}[narrativebox,title={Legal framework shields ministers from accountability}]
\begin{enumerate}[leftmargin=*, itemsep=1pt, topsep=1pt]
    \item Constitutional procedures preclude ministerial criminal liability for Tempi.
    \item Ministerial criminal liability in Tempi crash unsubstantiated.
    \item Government evades ministerial criminal liability for Tempi crash.
\end{enumerate}
\end{tcolorbox}

\begin{tcolorbox}[narrativebox,title={Opposition exploits the tragedy for political gain}]
\begin{enumerate}[leftmargin=*, itemsep=1pt, topsep=1pt]
    \item Opposition exploits tragedy for political gain.
    \item Opposition exploits tragedy to construct political narrative.
    \item Opposition exploits tragedy for political survival.
\end{enumerate}
\end{tcolorbox}

\begin{tcolorbox}[narrativebox,title={Railway privatization caused systemic safety failures}]
\begin{enumerate}[leftmargin=*, itemsep=1pt, topsep=1pt]
    \item Railway privatization caused systemic safety failures.
    \item Privatization caused unaccountable systemic railway safety failures.
    \item Railway privatization under liberalization caused systemic safety failures.
\end{enumerate}
\end{tcolorbox}

\end{minipage}

\caption{Representative narratives for the final narrative clusters (Table~\ref{tab:final_taxonomy}). For each narrative family, three illustrative examples are shown, selected using a combined criterion that rewards both silhouette-based centrality and lexical distinctiveness against all other clusters.}
\label{fig:distinctive_representatives}
\end{figure*}

\begin{algorithm*}[t]
\caption{The \textsc{CART} Framework}
\label{alg:pipeline}

\KwIn{Parliamentary corpus}
\KwOut{Narrative taxonomy}

\algstage{Stage I -- Thematic Filtering}

Encode every parliamentary paragraph using contextual embeddings\;

Reduce embedding dimensionality using PCA\;

\For{$k=15,\ldots,100$}{
    Cluster embeddings with k-means\;

    Compute the micro-averaged Silhouette score\;

    Compute the macro-averaged Silhouette score\;

    Compute
    \[
    H(k)=
    \frac{
        2S_{\mathrm{micro}}S_{\mathrm{macro}}
    }{
        S_{\mathrm{micro}}+S_{\mathrm{macro}}
    }
    \]
}

Select the value of $k$ maximizing $H(k)$\;

Cluster the complete paragraph corpus\;

Estimate assignment confidence using CAKE\;

Remove paragraphs belonging to the lowest confidence quartile\;

Repeat cluster selection on the confidence-filtered corpus\;

Construct the refined thematic partition\;

\algstage{Stage II -- Narrative Extraction}

For every refined cluster, construct metadata consisting of

\Indp
cluster size\;

five highest-confidence paragraphs\;

top TF--IDF terms\;
\Indm

Prompt the LLM to

\Indp
generate a descriptive cluster title\;

identify Tempi-related clusters\;
\Indm

Collect all paragraphs belonging to the selected clusters\;

\ForEach{paragraph}{
Generate one concise narrative statement\;

If the paragraph does not express a Tempi-related narrative,

discard it\;
}

Store the extracted narrative statements\;

\algstage{Stage III -- Narrative Cluster Analysis}

Encode narrative statements using contextual embeddings\;

Determine the optimal number of clusters using Composite Silhouette\;

Cluster the narrative embeddings using K-SIL\;

\ForEach{cluster}{

Retrieve representative high-Silhouette narrative statements\;

Compute contrastive TF--IDF terms\;

Assign an interpretable narrative-family label\;
}

Return the narrative taxonomy\;

\end{algorithm*}

\end{document}